\documentclass[10pt,twocolumn,letterpaper]{article}

\usepackage{iccv}
\usepackage{times}
\usepackage{epsfig}
\usepackage{graphicx}
\usepackage{amsmath}
\usepackage{amssymb}
\usepackage{multirow}
\usepackage{booktabs}
\usepackage{subfigure}
\usepackage{authblk}
\usepackage[breaklinks=true,bookmarks=false]{hyperref}
\newcommand{\redfont}[1]{{\color{red}{#1}}}
\newcommand{\bluefont}[1]{{\color{blue}{#1}}}

\iccvfinalcopy % *** Uncomment this line for the final submission

\def\iccvPaperID{****} % *** Enter the ICCV Paper ID here

\ificcvfinal\fi

\begin{document}

%%%%%%%%% TITLE
\title{Adaptive Reconstruction Network for Weakly Supervised \\ Referring Expression Grounding}

\author[1,2]{Xuejing Liu}
\author[1]{Liang Li\thanks{Corresponding author.}}
\author[1]{Shuhui Wang}
\author[3]{Zheng-Jun Zha}
\author[1,2]{Dechao Meng}
\author[2,1,4]{Qingming Huang}
\affil[1]{Key Lab of Intell. Info. Process., Inst. of Comput. Tech., CAS, Beijing, China}
\affil[2]{University of Chinese Academy of Sciences, Beijing, China}
\affil[3]{University of Science and Technology of China, Hefei, China}
\affil[4]{Peng Cheng Laboratory, Shenzhen, China}
\affil[ ]{\textit {{\{xuejing.liu, liang.li, dechao.meng\}@vipl.ict.ac.cn, wangshuhui@ict.ac.cn, zhazj@ustc.edu.cn, qmhuang@ucas.ac.cn}}}
%\and

\maketitle
\ificcvfinal\thispagestyle{empty}\fi

%%%%%%%%% ABSTRACT
\begin{abstract}
   Weakly supervised referring expression grounding aims at localizing the referential object in an image according to the linguistic query,
   %(language expressions)
   where the mapping between the referential object and query is unknown in the training stage.
   %Most recent work ignores the location and context information of the referential object, which has drawback that they cannot distinguish the  
   To address this problem, we propose a novel end-to-end adaptive reconstruction network (ARN). It builds the correspondence between image region proposal and query in an adaptive manner: adaptive grounding and collaborative reconstruction.
   Specifically, we first extract the subject, location and context features to represent the proposals and the query respectively. Then, we design the adaptive grounding module to compute the matching score between each proposal and query by a hierarchical attention model.
   Finally, based on attention score and proposal features, we reconstruct the input query with a collaborative loss of language reconstruction loss, adaptive reconstruction loss, and attribute classification loss.  This adaptive mechanism helps our model to alleviate the variance of different referring expressions.
   Experiments on four large-scale datasets show ARN outperforms existing state-of-the-art methods by a large margin.
   Qualitative results demonstrate that the proposed ARN can better handle the situation where multiple objects of a particular category situated together\footnote{Code is available at \url{https://github.com/GingL/ARN}}.
\end{abstract}

\section{Introduction}

%介绍任务及其应用.
Referring expression grounding (REG), also known as phrase localization, has been a surge of interest in both computer vision and natural language processing \cite{DBLP:conf/cvpr/MaoHTCY016, DBLP:conf/cvpr/HuXRFSD16, DBLP:conf/eccv/RohrbachRHDS16,DBLP:conf/cvpr/Yu0SYLBB18,8695120,DBLP:conf/mm/0003WJH18,ACMMM19YSJ,DBLP:journals/tip/WuWSH19}. Given a query (referring expression) in natural language and an image, REG is to find the corresponding location of the referential object. % in a free-form natural language query.
%For example, as Fig. \ref{pipeline} shows, given the query ``man in white on the left holding a bat'', REG should localize to the referential man (red bounding box) in the image. %加一个图，介绍具体一点
REG can be widely used in interactive applications, such as robotic navigation \cite{DBLP:conf/acl/ThomasonSM17, DBLP:conf/cvpr/AndersonWTB0S0G18}, visual Q\&A \cite{DBLP:conf/cvpr/DasDGLPB18, DBLP:conf/cvpr/GordonKRRFF18}, or photo editing \cite{DBLP:journals/tog/ChengZLVSCMT14}.
% component of connecting the humans。。。。
\begin{figure}
	\centering
	\includegraphics[width=8cm]{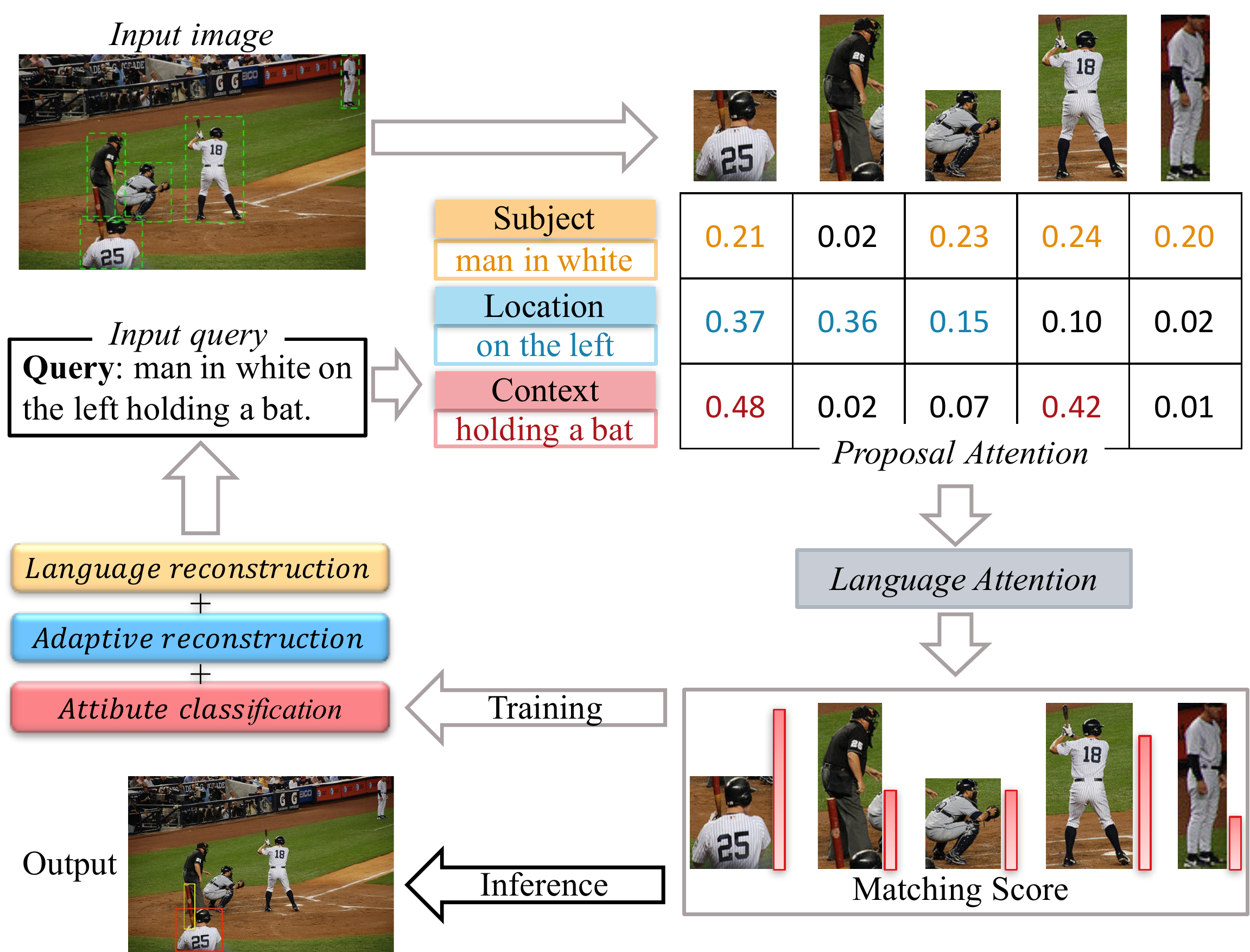}
	\caption{The proposed adaptive reconstruction network (ARN). Given a query and an image with region proposals, ARN localizes the referential object through adaptive grounding and collaborative reconstruction.}
	\label{pipeline}
\end{figure}

% 介绍弱监督方法 为什么要用弱监督？什么是弱监督？
% 当前方法大都有监督  可以分成first， second两点

Training the REG model in a supervised manner requires expensive annotated data that explicitly draw the connection between the input query and its corresponding object proposal in the image. Besides, limited to the training data, supervised REG models can only handle the grounding with certain categories, which cannot meet the demand for real-world applications. Here we focus on weakly supervised REG task, where only the image-query pairs are used for training without the mapping information between the query and the object proposal.

Previous weakly supervised methods \cite{DBLP:conf/eccv/RohrbachRHDS16,DBLP:conf/cvpr/ChenGN18,DBLP:conf/cvpr/ZhangNC18} learn to ground by reconstructing the input query.
Xiao {\it et al.}~\cite{DBLP:conf/cvpr/XiaoSL17} generate attention mask to localize linguistic query based on image-phrase pairs and language structure.
Zhao {\it et al.}~\cite{DBLP:conf/cvpr/0006LZF18} try to find the location of the referential object by searching over the entire image.
The above methods only exploit the visual appearance features of proposals during grounding and reconstruction. However, they ignore the discriminative information of location and context from the referential object, and cannot distinguish a specific object where multiple objects of a particular category situated together.
%For the query in Fig. \ref{pipeline}, not only the subject (``man in white''), but location (\eg, ``on the left'') and context (e.g., ``holding a bat'') play an essential part in distinguishing the referential object as well.
As shown in Fig. \ref{pipeline} , given the query of ``man in white on the left holding a bat'', apart from the subject (``man in white''), location (``on the left'') and context (``holding a bat'') play an essential part in distinguishing the referential object.

Recently Yu {\it et al}.~\cite{DBLP:conf/cvpr/Yu0SYLBB18} find that people tend to use different syntax structures when referring to an object, and this brings the variance of different referring expressions.
Taking Fig. \ref{pipeline} as an example, if the query is only ``man in black'', it can be grounded using subject features only. Similarly, ``man on the far right'' concentrates more on location features, and ``man on the right of the man in black'' should focus more on context features.
Therefore, the grounding is triggered based on what features are present in the referring expression.

In light of these observations, we propose a novel end-to-end weakly supervised REG method, coined Adaptive Reconstruction Network (ARN).
It learns the mapping between image region proposal and query upon the subject, location and context information in an adaptive manner.
Fig. \ref{pipeline} shows the pipeline of ARN, that consists of two modules: adaptive grounding, and collaborative reconstruction.

\textbf{Adaptive Grounding.} First, we extract the subject, location, and context features of both the query and each region proposal in an image.
Specifically, for the query, we introduce a recurrent net to parse it into these three features.
For a proposal, we extract its visual appearance feature as the subject feature by Faster R-CNN \cite{DBLP:conf/nips/RenHGS15}. Moreover, the location feature of the proposal consists of absolute position and relative locations with other proposals of the same category in the image. Furthermore, the context feature of the proposal is represented by concatenating the visual and relative location features of its surrounding proposals.
%For a proposal, the subject feature is its visual appearance feature extracted as the subject feature by Faster R-CNN \cite{DBLP:conf/nips/RenHGS15};
%its location feature consists of absolute position and relative locations with other proposals of the same category in the image;
%we concatenate the visual and relative location features of its surrounding proposals as its context feature, which model the relationship between the proposal and its environment.
%For a proposal, we extract its visual appearance feature as the subject feature by Faster R-CNN \cite{DBLP:conf/nips/RenHGS15};
%its location feature consists of absolute position and relative locations with other proposals of the same category in the image;
%we concatenate the visual and relative location features of its surrounding proposals as its context feature, which model the relationship between the proposal and its environment.
Second, we propose the adaptive grounding module to compute the matching score between each proposal and query by a hierarchical attention model.
The first attention helps generate attention scores upon subject, location, and context for each proposal respectively. The second one further learns the attention score of the above three components based on the syntax structure of the query. This module can alleviate the variance of different referring expressions.

\textbf{Collaborative Reconstruction.} We design a collaborative loss to better formulate the measurement of weakly supervised grounding. The loss function derives from the following three parts.
\textit{Language reconstruction} directly reconstructs the input query based on the attentive proposal features. \textit{Adaptive reconstruction} reconstructs attentive hidden features of subject, location and context respectively.
\textit{Attribute classification} leverages the attribute information of candidate proposal upon the subject to improve the grounding ability.

Both modules of ARN can be trained in an end-to-end manner. At the inference stage, ARN only utilizes the adaptive grounding to localize the referent without reconstruction.
To summary, the main contribution of this paper is three-fold:
%Our contributions can be summarized as follows.
\begin{itemize}
	\item We propose an end-to-end adaptive reconstruction network that models the mapping between input query and image upon subject, location and context features. ARN adaptively grounds the candidate proposals by hierarchical attention, which could alleviate the variance of different referring expressions.
	
	%In light of these observations, we propose a novel end-to-end weakly supervised REG method,    coined Adaptive Reconstruction Network (ARN).
	%It learns the mapping between image region proposal and query upon the subject, location and context information in an adaptive manner.
	%Fig. \ref{pipeline} shows the architecture of ARN, that consists of two modules: adaptive grounding, and collaborative reconstruction.
	%better formulate the measurement of weakly supervised grounding
	
	\item We design a collaborative reconstruction module to reconstruct the input query based on the matching score and proposal features. A collaborative loss of language reconstruction, adaptive reconstruction, and attribute classification are formulated for the measurement of adaptive grounding.
	
	\item Comparison experiments on the RefCLEF and three MS-COCO datasets show that the proposed ARN achieves state-of-the-art results in the weakly supervised REG task.
\end{itemize}
%------------------------------------------------------------------------
\begin{figure*}[ht]
	\centering
	\subfigure[Visual feature encoding.]{
		\label{vis_feats}
		\includegraphics[height=5cm]{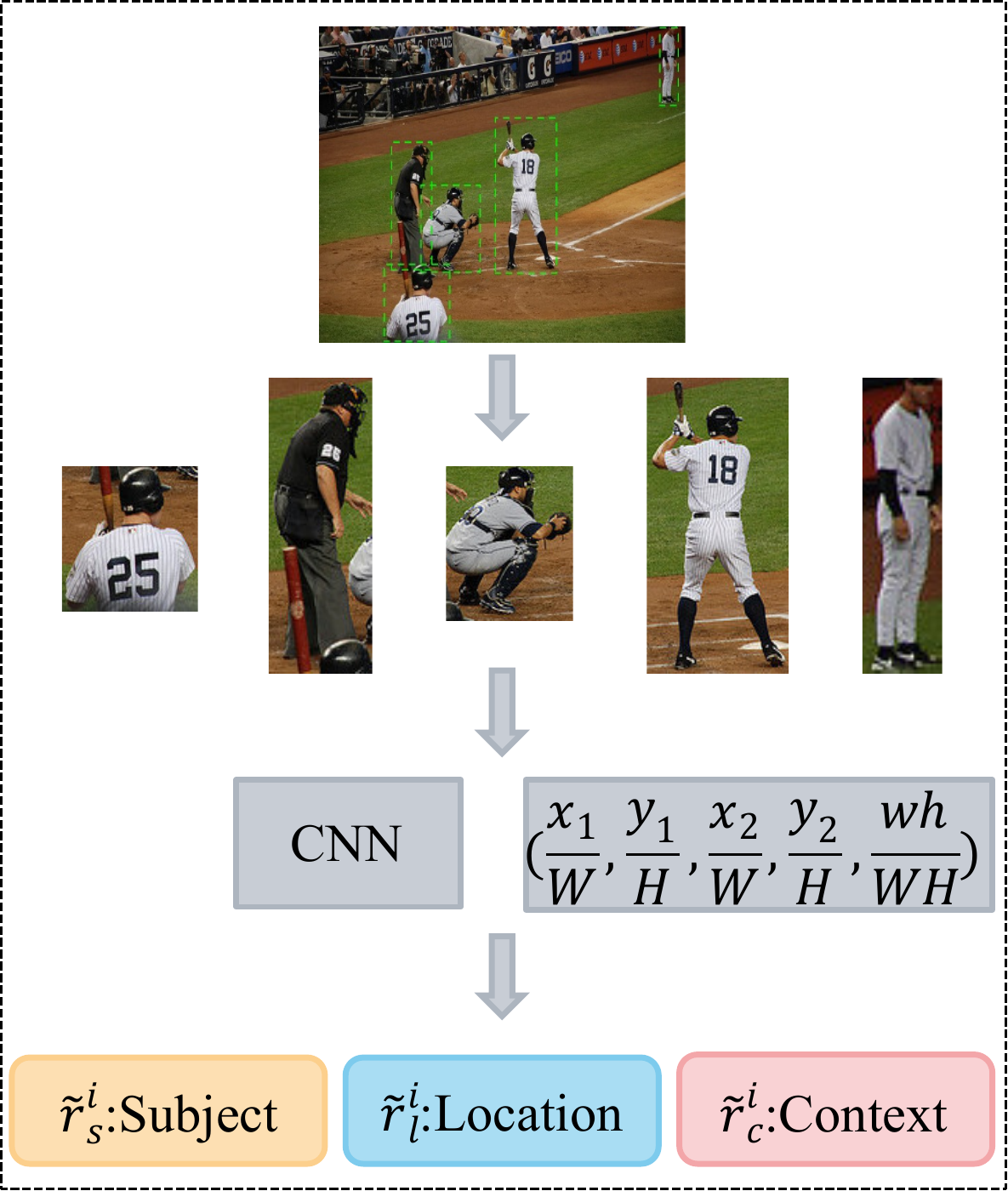}}
	%\hspace{0.22cm}
	\subfigure[Language feature encoding.]{
		\label{lan_feats}
		\includegraphics[height=5cm]{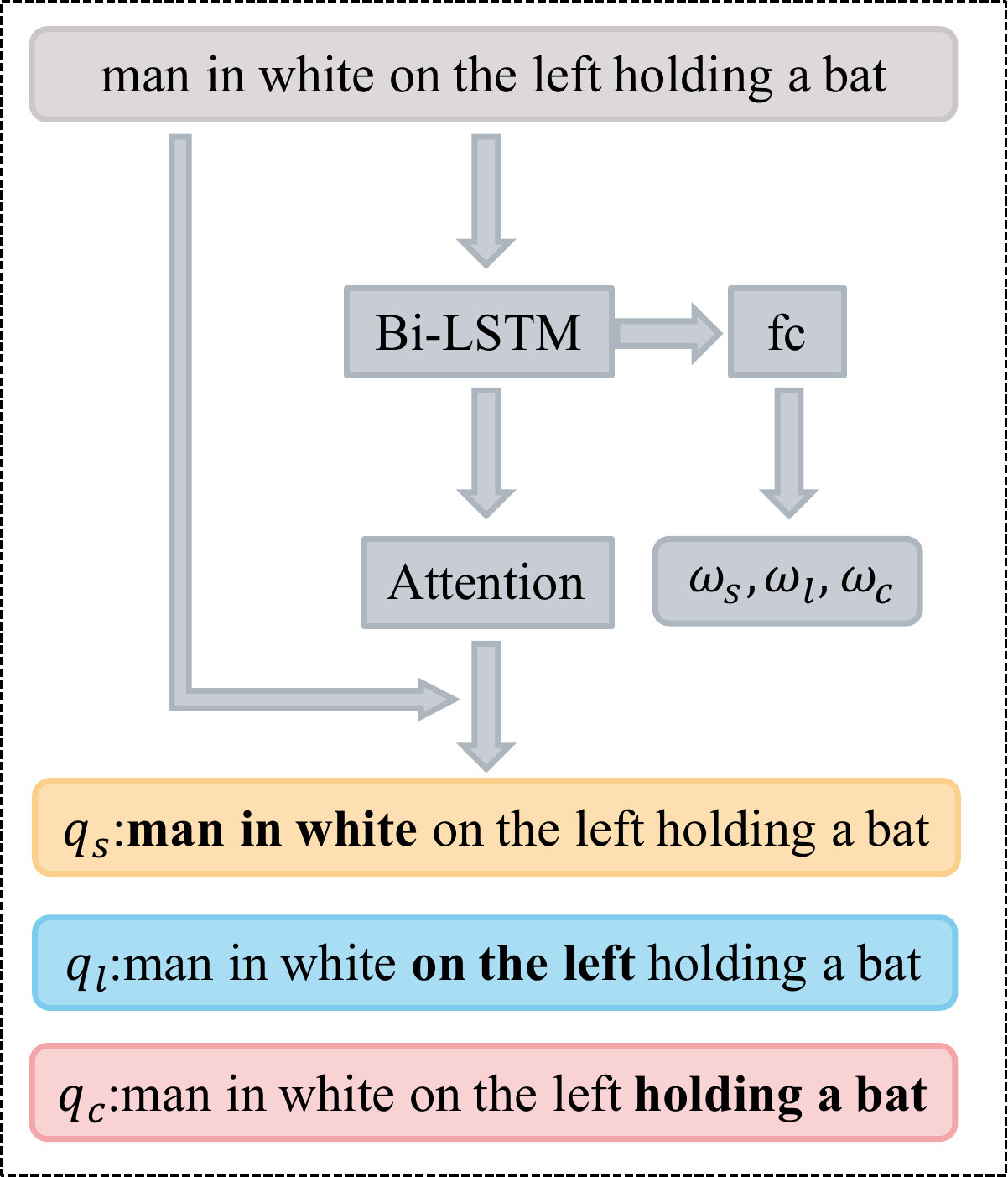}}
	%\hspace{0.22cm}
	\subfigure[Adaptive grounding.]{
		\label{grounding}
		\includegraphics[height=5cm]{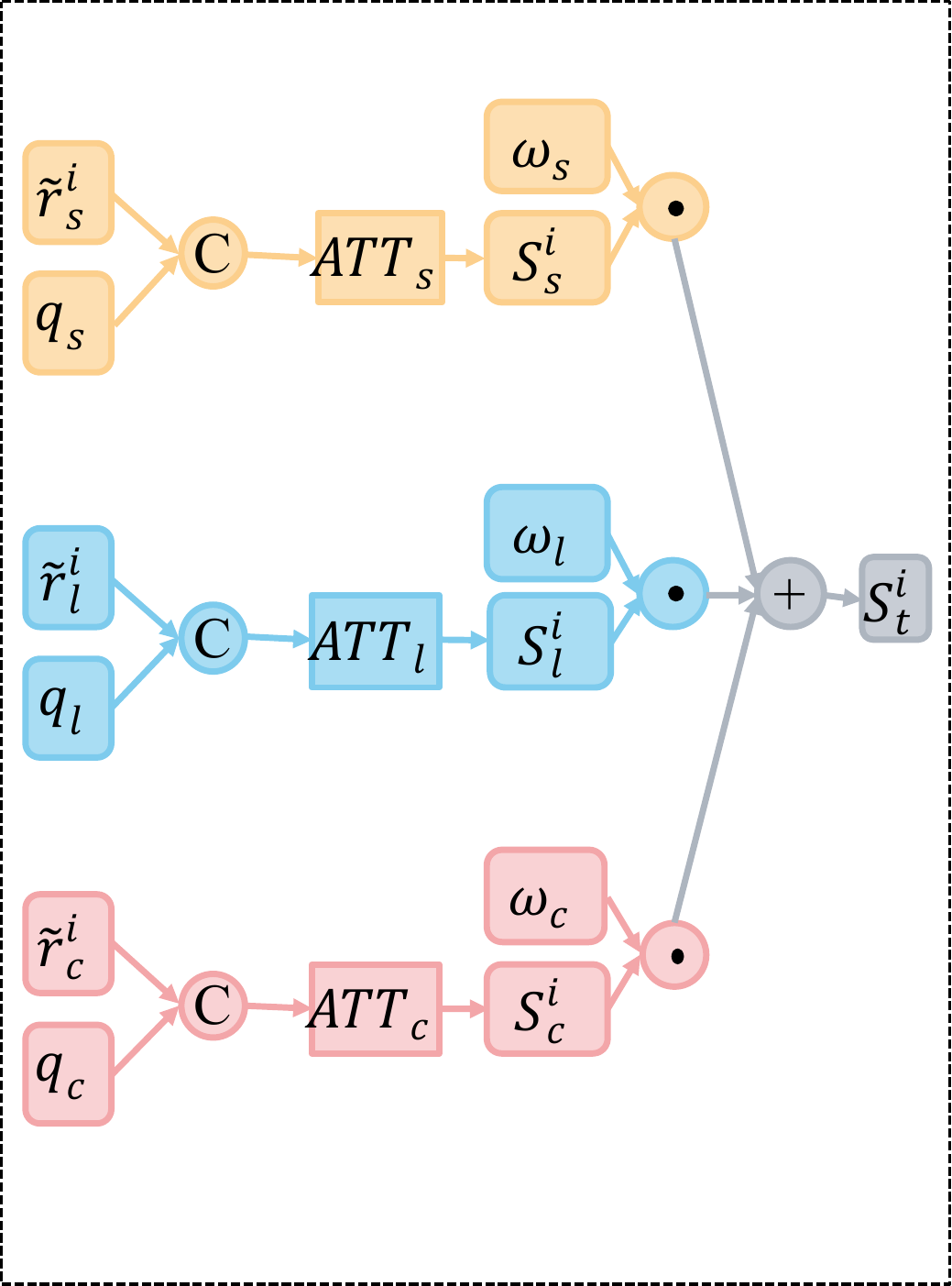}}
	%\hspace{0.22cm}
	\subfigure[Collaborative Reconstruction.]{
		\label{reconstruct}
		\includegraphics[height=5cm]{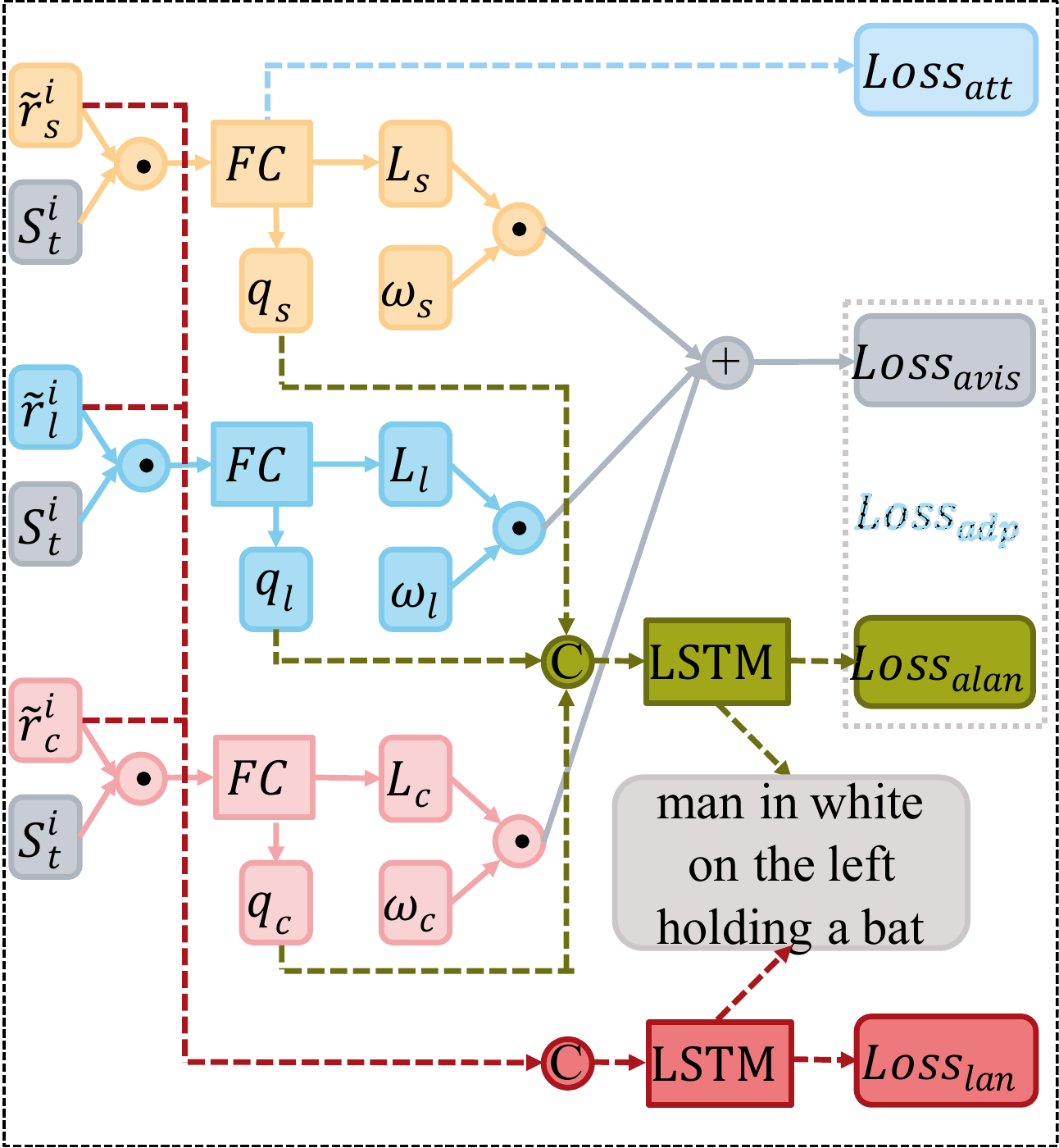}}
	\caption{The network architecture of the proposed ARN. It consists of feature encoding (Section \ref{feats_enc}), adaptive grounding (Section \ref{adp_loc}) and collaborative reconstruction (Section \ref{adp_rec}). Collaborative reconstruction module contains three losses: attribute classification loss, adaptive reconstruction loss (including adaptive language reconstruction and adaptive visual reconstruction loss) and language reconstruction loss.
		Visual features are pre-extracted from external networks. Language feature encoding, adaptive grounding and collaborative reconstruction are trained as an end-to-end network. The reconstruction module is not needed during inference.  ATT: attention layer. $\oplus$: plus operation. $\odot$: element-wise vector multiplication. C: vector concatenation.}
	\label{model}
	\vspace{1.2em}
\end{figure*}

\section{Related Work}

%什么叫grounding？有什么区别？当前有哪两类方法？每类方法具体是怎么做的？
\noindent \textbf{Referring Expression Grounding (REG).} REG \cite{DBLP:conf/emnlp/KazemzadehOMB14, DBLP:conf/naacl/MitchellDR13, DBLP:conf/emnlp/FitzGeraldAZ13, DBLP:conf/cvpr/MaoHTCY016, DBLP:conf/eccv/YuPYBB16,  DBLP:conf/cvpr/YuTBB17, DBLP:journals/corr/abs-1812-03426,ACMMM19LXJ} is also known as referring expression comprehension or phrase localization, which is the inverse task of referring expression generation.
%问题描述
REG aims to localize the corresponding object described by a free-form natural language query in an image.
%This problem can be formulated as follows. 
Given an image $I$, a query $q$ and a set of region proposals $\{r_i\}^N_{i=1}$, REG selects the best-matched region $r^*$ according to the query.
%有哪些数据集？
%After the release of standard datasets \cite{DBLP:conf/emnlp/KazemzadehOMB14, DBLP:conf/eccv/YuPYBB16, DBLP:conf/cvpr/MaoHTCY016, DBLP:journals/ijcv/PlummerWCCHL17}, REG has attracted great interests.
%方法分类
Most REG methods can be roughly divided into two kinds.
One is CNN-LSTM based encoder-decoder structure to model $P(q|I,r)$ \cite{DBLP:conf/cvpr/MaoHTCY016, DBLP:conf/eccv/YuPYBB16, DBLP:conf/eccv/NagarajaMD16, DBLP:conf/cvpr/HuXRFSD16, DBLP:conf/cvpr/LuoS17,DBLP:journals/ieeemm/LiJZWH13,DBLP:journals/tip/SongWHT17}.
The other is the joint vision-language embedding framework to model $P(q, r)$.
During training, the supervision is object proposal and referring expression pairs $(r_i, q_i)$ \cite{DBLP:conf/eccv/RohrbachRHDS16, DBLP:conf/cvpr/WangLL16, DBLP:conf/iccv/Liu0017, DBLP:conf/iccv/ChenKN17, DBLP:conf/cvpr/Yu0SYLBB18,DBLP:journals/tmm/LiJH12,DBLP:conf/mm/WangCZH018}.
Recently, MattNet \cite{DBLP:conf/cvpr/Yu0SYLBB18} adopts subject, location and relation features on supervised REG and gets state-of-the-art results.
The above features prove to be effective in the grounding task, which are also used as the original feature representation in our method. But we design the collaborative reconstruction to bridge the gap between supervised and weakly supervised learning, achieving impressive results on weakly supervised REG.\\

%Understanding the given natural language and grounding the referred objects are fundamental steps for many interactive tasks.

\iffalse
% 介绍了一部分referring expression以及啥叫generation
\textbf{Referring expression} is related with two different tasks, generation and comprehension. Referring expression generation \cite{DBLP:conf/emnlp/KazemzadehOMB14} is a special case of image caption \cite{DBLP:conf/eccv/FarhadiHSYRHF10,DBLP:conf/cvpr/VinyalsTBE15,DBLP:conf/inlg/MitchellDR10,DBLP:conf/naacl/MitchellDR13}, which is to generate an sentence of arbitrary length to describe the general content of an image. Referring expression generation focus more on the specific object in an image.
\fi

%无监督的方法

\noindent \textbf{Weakly Supervised Referring Expression Grounding.} Weakly supervised REG only has image-level correspondence, and there is no mapping between image regions and referring expressions. To solve this problem,  Rohrbach {\it et al.}~\cite{DBLP:conf/eccv/RohrbachRHDS16} propose a framework which learns to ground by reconstructing the given referring expression through attention mechanism. Based on this framework, Chen {\it et al.}~\cite{DBLP:conf/cvpr/ChenGN18} design knowledge aided consistency network, which reconstructs both the input query and proposal's information. 
%They also introduce a Knowledge-based pooling gate to filter out the unrelated regions. 
Xiao {\it et al.}~\cite{DBLP:conf/cvpr/XiaoSL17} ground arbitrary linguistic phrase in the form of spatial attention mask and propose a network with discriminative and structural loss.
%One is a standard discriminative loss to match the corresponding image-phrase pair. The other is a structural loss to make use of rich structure information of language.
Different from selecting the optimal region from a set of region proposals, Zhao {\it et al.}~\cite{DBLP:conf/cvpr/0006LZF18} propose multi-scale anchored transformer network, which can search the entire spatial feature map by taking region proposals as anchors to get more accurate location. Zhang {\it et al.}~\cite{DBLP:conf/cvpr/ZhangNC18} propose a variational Bayesian method to exploit the relationship between the referent and context. %Their method can significantly reduce the search space of context.

% 如果字数不够还可以加一节： visual and language，介绍与grounding相关的应用，比如caption，人机交互等，介绍grounding可以用在哪些实际生活场景中，体现其重要意义
%\textbf{Visual and Language.}
%Referring expression grounding is the fundamental task for many complex tasks, such as robotic navigation \cite{DBLP:conf/cvpr/AndersonWTB0S0G18}, visual Q\&A \cite{DBLP:conf/cvpr/DasDGLPB18, DBLP:conf/cvpr/GordonKRRFF18}, and visual chatbot \cite{DBLP:conf/iccv/DasKMLB17}. \cite{DBLP:conf/cvpr/DasDGLPB18} present a new task, where an agent is asked a question in a 3D environment. The agent will navigate to explore the environment and answer this question. \cite{DBLP:conf/cvpr/AndersonWTB0S0G18} provide the first dataset for visually-grounded natural language navigation in real buildings. They also design the Matterport3D Simulator, which is a large-scale reinforcement learning environment based on real imagery.  The simulator interprets a natural-language navigation instruction based on its environment. \cite{DBLP:conf/cvpr/GordonKRRFF18} introduce a new task called interactive question answer, where an agent is asked to answer a question related to its environment.
%Understanding the given natural language and grounding the referred objects are fundamental steps for the above challenging and realistic tasks. % 形容重要且基础用什么词？？？ 表示实用和复杂，是不是可以用更好的词？？

\section{Methodology}

We propose an adaptive reconstruction network (ARN) to ground the target proposal described by the query in weakly supervised scenario, where the training data do not have the region-query correspondence.
This problem can be formulated as follows. Given an image $I$, a query $q$ and a set of region proposals $\{r_i\}^N_{i=1}$, we aim at selecting the best-matched region $r^*$ according to the query without knowing any $(q, r_i)$ pair.
% 需要大致介绍一下自己的方法，网络结构是怎样的
%Previous methods usually utilize attention mechanism to choose the most probable region to reconstruct its corresponding sentence or phrase. We take advantage of the same reconstruction mechanism.
ARN chooses the most probable proposal through adaptive grounding, then reconstructs its corresponding query with a collaborative loss.  The whole network architecture is shown in Fig. \ref{model}.

\subsection{Feature Encoding}
\label{feats_enc}
%解释一下自己在视觉特征中不再使用global features
%In this subsection, we introduce the visual and language features extraction in detail.
%Visual features are pre-extracted from external networks or arithmetic operators for region proposals. 
%Global visual features are not utilized in our method, as global features will introduce some ambiguity to the grounding task \cite{DBLP:conf/eccv/YuPYBB16}. 
%The language features are extracted through LSTM \cite{DBLP:journals/neco/HochreiterS97} and attention mechanism \cite{DBLP:conf/nips/VaswaniSPUJGKP17}. %如果之后有时间尝试external language parser，这可以再提一下。

\subsubsection{RoI Features}
% 视觉特征分三种，每种各是什么及其表示
For each object proposal $r_i$, the subject, location and context features are extracted as shown in Fig. \ref{vis_feats}.

\textbf{Subject feature} is extracted as visual appearance features of proposals.
% Faster R-CNN \cite{DBLP:conf/nips/RenHGS15} is used as the backbone network to extract the region features.
We run the forward propagation of Faster R-CNN based on ResNet \cite{DBLP:conf/cvpr/HeZRS16} for each image, and crop its C3 and C4 features as the subject feature $\widetilde { r } _ { s } ^ { i } = f_{CNN}(r_i)$. The C3 features represent lower-level features such as colors and shapes while C4 features contain higher-level representations. %有没有更好的词？？？

\textbf{Location feature} consists of absolute position and relative locations with other objects of the same category in the image. Following \cite{DBLP:conf/eccv/YuPYBB16, DBLP:conf/cvpr/YuTBB17, DBLP:conf/cvpr/Yu0SYLBB18}, the absolute location feature of each proposal is decoded as a 5-dim vector $r_l^i = \left[ \frac { x _ { t l } } { W } , \frac { y _ { t l } } { H } , \frac { x _ { b r } } { W } , \frac { y _ { b r } } { H } , \frac { w \cdot h } { W \cdot H } \right]$, denoting the top-left, bottom-right position and relative area of the proposal to the whole image. The relative location feature indicates the relative location information between the proposal and 5 surrounding proposals of the same category. 
For each surrounding proposal, we calculate its offset and area ratio to the candidate: $\delta r_l^{ij} = \left[ \frac { \left[ \Delta x _ { t l } \right] _ { i j } } { w _ { i } } , \frac { \left[ \triangle y _ { t l } \right] _ { i j } } { h _ { i } } , \frac { \left[ \triangle x _ { b r } \right] _ { i j } } { w _ { i } } , \frac { \left[ \triangle y _ { b r } \right] _ { i j } } { h _ { i } } , \frac { w _ { j } h _ { j } } { w _ { i } h _ { i } } \right]$.
Finally, we concatenate the above absolute and relative location feature into the location feature of the proposal, which is a 30-dim vector: $\widetilde { r } _ { l } ^ { i } = \left[ r _ { l }^{i} ; \delta r _ { l }^{i} \right] $.

% 是否需要举些例子更清楚的表述为什么分成这三种特征？是否需要举些实例来讲清楚各部分到底对应的是什么？

\textbf{Context feature} represents the relationship between the candidate proposal and environment.
%, which is an important cue to localize the referential object. For example, ``the man holding a bat''.
Following \cite{DBLP:conf/cvpr/Yu0SYLBB18}, we choose 5 surrounding proposals as the relative ones for each proposal.
The feature of each proposal is composed of C4 feature $v_{ij}=f_{CNN}(r_{j})$ and its relative location feature. The relative location feature is encoded as follows: $\delta m _ { i j } = \left[ \frac { \left[ \triangle x _ { t l } \right] _ { i j } } { w _ { i } } , \frac { \left[ \triangle y _ { t l } \right] _ { i j } } { h _ { i } } , \frac { \left[ \triangle x _ { b r } \right] _ { i j } } { w _ { i } } , \frac { \left[ \triangle y _ { b r } \right] _ { i j } } { h _ { i } } , \frac { w _ { j } h _ { j } } { w _ { i } h _ { i } } \right]$. The context feature is $\widetilde { r } _ { c } ^ { ij } =  \left[ v _ { i j } ; \delta m _ { i j } \right]$. From above 5 proposals, we choose the one with the maximum response to the query as the final relative object,  denoted as $\widetilde { r } _ { c } ^ { i }$ .
%To further increase the expressive power, similar to [6, 12, 13, 27, 44, 45], we also embed the spatial representations of the proposal regions.
\subsubsection{Referring Expression Features}
\label{reffeats}
% 首先描述分成三部分，按顺序通过word embedding， LSTM。 之后又通过两种attetion， 一种生成weight， 一种生成subject， location，context的向量表示。weight的作用解释一下。公式化说明整个流程，不要忘记指向Fig
Corresponding to RoI features, the query features are also separated into \textbf{subject }$q_s$, \textbf{location} $q_l$ and \textbf{context} $q_c$ through attention mechanism, as shown in Fig. \ref{lan_feats}. Given an query $q = \left\{ w _ { t } \right\} _ { t = 1 } ^ { T }$,  first each word in $q$ is one-hot encoded and mapped into a word embedding $e_t$. Then the word embedding $e_t$ is fed into a bi-directional LSTM. The final representation $h_t = [\overrightarrow{h}_t, \overleftarrow{h}_t]$ is the concatenation of the hidden vectors in both directions. Words are attended in each query for better representation of subject, location and context through attention mechanism. Take subject feature $q_s$ as an example, its final hidden representation is calculated as follows:

\begin{equation}
\begin{aligned}
{ m } _ { t } &= \mathrm { fc } \left(  { h } _ { t } \right) ,\\
\alpha _ { t } &= \operatorname { softmax } _ { t } \left( { m } _ { t } \right) ,\\
{ q_s } &= \sum _ { t } \alpha _ { t }  { e } _ { t }.
\end{aligned}
\label{att_sub}
\end{equation}

Location feature $q_l$ and context feature $q_c$ can be obtained using the same mechanism.
Besides, three different weights upon subject, location and context are calculated from the hidden state vector of the bi-directional LSTM.
\begin{equation}
\left[ w _ { s } , w _ { l  } , w _ { c } \right] = \operatorname { softmax }_{w} \left( \mathrm { fc } \left( \left[{ h } _ { 0 }, { h } _ { T } \right] \right) \right)
\end{equation}

\subsection{Adaptive Grounding}
\label{adp_loc}
Based on the subject, location and context features of both the proposal and query, ARN localizes the query through a hierarchical attention model. The first attention is the \textbf{proposal attention}, which calculates the matching score between the proposals and query upon subject, location and context respectively. The second attention is \textbf{language attention}, which assigns different weights to subject, location and context based on the query to alleviate variance in queries.

Detailed process can be seen in Fig. \ref{grounding}, $\widetilde { r } _ { s } ^ { i }$, $\widetilde { r } _ { l } ^ { i }$ and $\widetilde { r } _ { c } ^ { i }$ are the visual features extracted from the region proposals in the image through CNN. $q_s$, $q_l$ and $q_c$ represent the language feature extracted from the query through bi-directional LSTM. Taking the subject as an example, $\widetilde { r } _ { s } ^ { i }$ and $q_s$ are first concatenated into one vector. Then the vector is fed into the proposal attention, which is a two layer perceptron, to get the corresponding matching score. The biases are omitted in Eq. (\ref{subscore}).
\begin{equation}
\label{subscore}
{  \overline{ s } } _ { x }^{i} = f _ { A T T } \left( q_x , \widetilde { r } _ { x } ^ { i } \right) = W _ { 2 } \phi_{\rm ReLU} \left( W _ { 1 } [ q_x , \widetilde { r } _ { x } ^ { i }] \right), x \in (s, l, c)
\end{equation}
\iffalse
\begin{equation}
\label{subscore}
\begin{aligned}
{ \overline{ s } } _ { s }^{i} &= f _ { A T T } \left( q_s , \widetilde { r } _ { s } ^ { i } \right) &= W _ { 2 } \phi_{\rm ReLU} \left( W _ { 1 } [ q_s , \widetilde { r } _ { s } ^ { i }] \right) \\
{  \overline{ s } } _ { l }^{i} &= f _ { A T T } \left( q_l , \widetilde { r } _ { l } ^ { i } \right) &= W _ { 2 } \phi_{\rm ReLU} \left( W _ { 1 } [ q_l , \widetilde { r } _ { l } ^ { i }] \right) \\
{  \overline{ s } } _ { c }^{i} &= f _ { A T T } \left( q_c , \widetilde { r } _ { c } ^ { i } \right) &= W _ { 2 } \phi_{\rm ReLU} \left( W _ { 1 } [ q_c , \widetilde { r } _ { c } ^ { i }] \right)
\end{aligned}
\end{equation}
\fi

We normalize the scores using softmax.
\begin{equation}
{ s } _ { x } ^ { i }  = \operatorname { softmax } _ { i } \left( \overline { s } _ { x } ^ { i } \right), \quad x \in (s, l, c) 
\end{equation}
\iffalse
\begin{equation}
\begin{aligned}  { s } _ { s } ^ { i } & = \operatorname { softmax } _ { i } \left( \overline { s } _ { s } ^ { i } \right) \\ { s } _ { l } ^ { i } & = \operatorname { softmax } _ { i } \left( \overline { s } _ { l } ^ { i } \right) \\
{ s } _ { c } ^ { i } & = \operatorname { softmax } _ { i } \left( \overline { s } _ { c } ^ { i } \right) \end{aligned}
\end{equation}
\fi
The total score is calculated based on language attention, which is the linear combination of the three sub-score.
The final score represents the probability of region \textit{i} matching query \textit{q} considering subject, location and context.
The weights are calculated based on the query.
%The weights are calculated based on the referring expressions, as subsection \ref{reffeats} shows.
%which is the linear combination of the three sub-score,
\begin{equation}
S_t^i = w _ { s }s _ { s } ^ { i } + w _ { l } s _ { l } ^ { i } + w _ { c } s _ { c } ^ { i }
\end{equation}

\subsection{Collaborative Reconstruction}
\label{adp_rec}

\begin{figure}
	\centering
	\subfigure[]{
		\label{adaploss}
		\includegraphics[height=5cm]{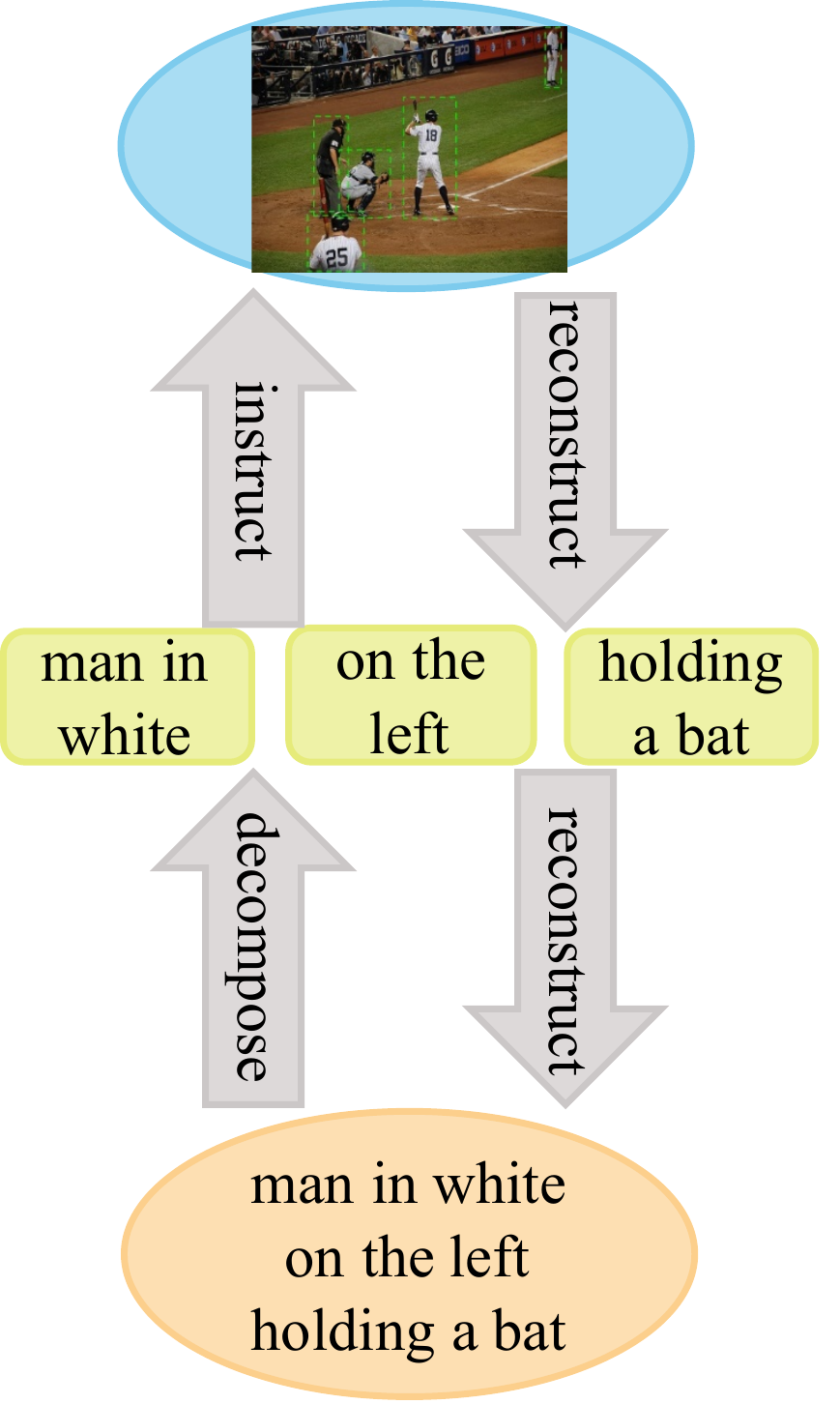}
	}
	\hspace{0.2 cm}
	\subfigure[]{
		\label{langloss}
		\includegraphics[height=5cm]{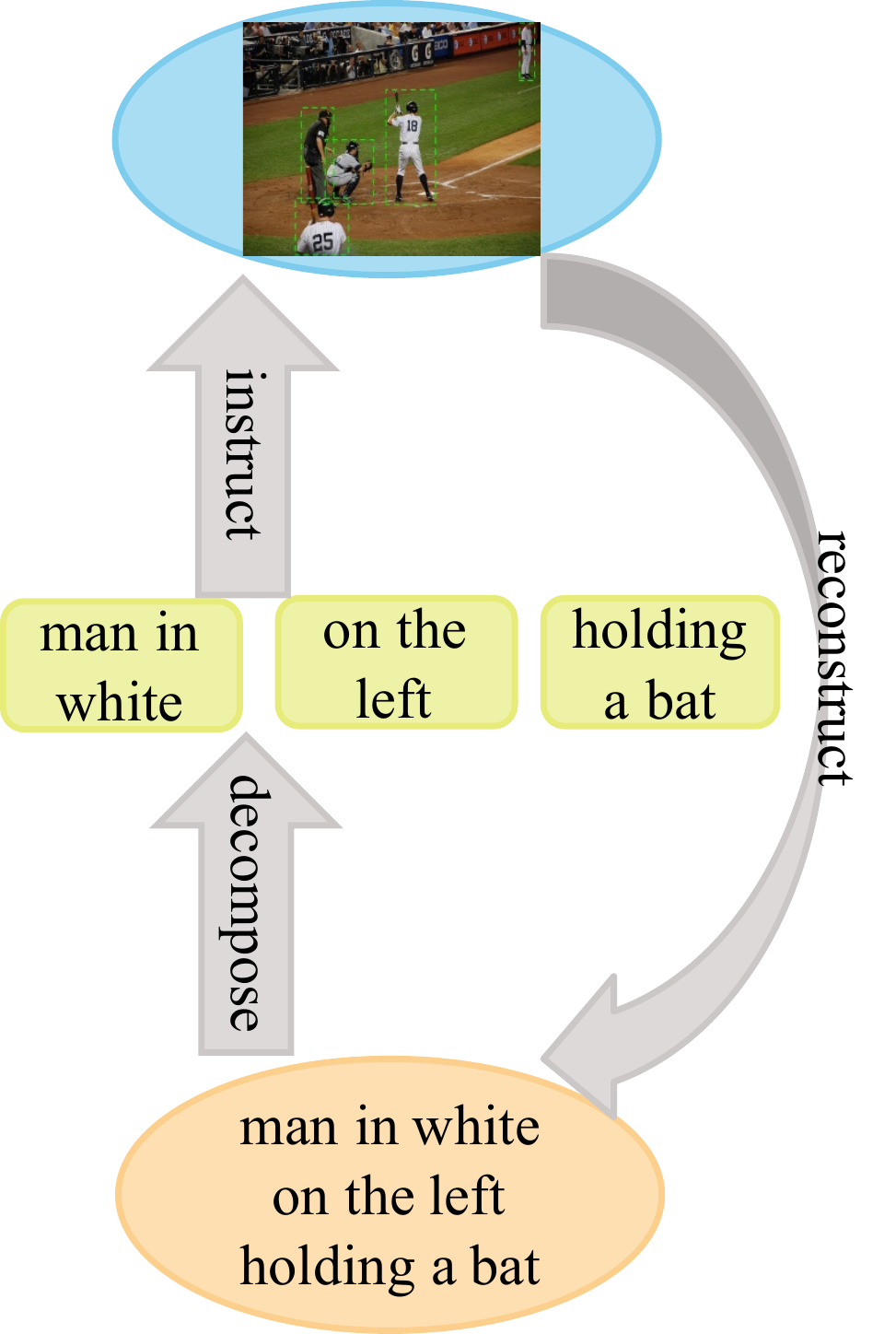}}
	\caption{The sketch map of (a) Adaptive reconstruction and (b) Language reconstruction.}
	\label{RecLoss}
\end{figure}
Because there are no mapping data between the query and the proposal of the image in the weak supervised training stage, 
the collaborative reconstruction is used to formulate the measurement of the grounding.
The collaborative loss is designed with three losses, as shown in Fig.~\ref{reconstruct}. Adaptive reconstruction reconstructs attentive hidden features of subject, location and context respectively. Language reconstruction directly reconstructs the input query based on the attentive features of proposals.  Attribute classification takes advantage of attribute information of the referential proposal.

\subsubsection{Adaptive Reconstruction Loss}
Adaptive reconstruction loss utilizes three hidden vectors (subject, location, context) to bridge the gap between the input query and proposals, as Fig. \ref{adaploss} shows. This loss consists of two sub-losses, adaptive visual reconstruction loss and adaptive language reconstruction loss.
This is inspired by the idea that reconstructing different linguistic query using corresponding features of proposals can better handle the variance among different expressions in the datasets.
The adaptive visual reconstruction loss is to reconstruct query features $q _ { s }$, $q _ { l }$ and $q _ { c }$ using features of proposals $\widetilde { r } _ { s } ^ { i }$, $\widetilde { r } _ { l } ^ { i }$ and $\widetilde { r } _ { c } ^ { i }$. We first compute a weighted sum over the different visual features and the matching scores.
%计算attention的vis feature
\begin{equation}
\tilde {v} _ { s } = \sum _ { i = 1 } ^ { N } S _ { t } ^ { i } \tilde { r } _ { s } ^ { i },\quad
\tilde {v} _ { l } = \sum _ { i = 1 } ^ { N } S _ { t } ^ { i } \tilde { r } _ { l } ^ { i },\quad
\tilde {v} _ { c } = \sum _ { i = 1 } ^ { N } S _ { t } ^ { i } \tilde { r } _ { c } ^ { i }
\end{equation}
%通过FC层使lan和vis feats维度相同
The aggregation of the proposal features from the attentive proposals are then fed into a fully connected layer to get the same dimension with the language features. %感觉get the same dimension有点不太对
\begin{equation}
v _ { s } = FC(\tilde {v} _ { s }),\quad
v _ { l } = FC(\tilde {v} _ { l }),\quad
v _ { c } = FC(\tilde {v} _ { c })
\end{equation}
%计算sub vis loss for each module，使用mse
% referring expression和query可不可以混用啊？？？
Then we use the attentive proposal features $v_s$, $v_l$ and $v_c$ to reconstruct the language features $q _ { s }$, $q _ { l }$ and $q _ { c }$ extracted from the original query. We use Mean Squared Error (MSE) criterion to minimize the distance between the proposal features and language features.
\begin{equation}
L_x ={\rm MSE}(v _ { x }, q_x), \quad x \in (s, l, c)
\end{equation}

%\begin{equation}
%\begin{aligned}
%L_s &={\rm MSE}(v _ { s }, q_s),\\
%L_l &= {\rm MSE}({v} _ { l }, q_l),\\
%L_c &= {\rm MSE}({v} _ { c }, q_c)
%\end{aligned}
%\end{equation}
%计算总的vis loss
The final adaptive visual reconstruction loss is the weighted sum of the subject reconstruction loss, location reconstruction loss and context reconstruction loss. The weights are calculated based on the query, as subsection \ref{reffeats} shows.
\begin{equation}
Loss_{avis} = w _ { s } L_s + w _ { l } L_l + w _ { c } L_c
\end{equation}
% 首先解释为什么需要language reconstruction部分
% language reconstruction部分，首先concat之后放入fc层再通过relu\
However, the language feature extraction network is trained together with the grounding and reconstruction network. To reach convergence as soon as possible,  network parameters might be set to zero roughly so that the network can not learn the correspondence between the visual modality and language modality. %是不是有更专业的词代替correspondence
To avoid this circumstance, we add an adaptive language reconstruction loss, which utilizes the language features $q _ { s }$, $q _ { l }$ and $q _ { c }$ to reconstruct the original query. %circumstance合适不合适？？？
First we concatenate $q _ { s }$ , $q _ { l }$ and $q _ { c }$, then feed it into a one-layer perceptron.
\begin{equation}
f_{alan}=\phi _ {\rm R e L U } \left(W_l( \left[ q _ { s } , q_l, q_c \right])+b_l  \right)
\end{equation}
%通过LSTM生成phrase
Based on the fused language features $f_{lan}$, we reconstruct the input query through LSTM. This is inspired by the query generation methods \cite{DBLP:conf/cvpr/DonahueHGRVDS15, DBLP:conf/cvpr/VinyalsTBE15}. The language features $f_{lan}$ are fed into a one-layer LSTM only at the first time step.
\begin{equation}
P ( q | f_{alan} ) = f _ {\rm L S T M } \left(f_{alan} \right)
\end{equation}
% 与原始的query进行对比
The language reconstruction network aims to maximize the likelihood of the ground-truth query $\hat { q }$ generated by LSTM, as Eq. (\ref{lanloss}) shows. B is the batch size.
\begin{equation}
\label{lanloss}
Loss _ { alan } = - \frac { 1 } { B } \sum _ { b = 1 } ^ { B } \log ( P ( \hat { q } | f_{alan} ) )
\end{equation}
%总的重建损失
The final adaptive reconstruction loss is the weighted sum of the language reconstruction loss and visual reconstruction loss. $\alpha$ and $\beta$ is the hyper-parameters defining the proportion of the two losses. In this adaptive reconstruction loss, both of the language and visual reconstruction loss are indispensable. 
\begin{equation}
Loss_{adp} = \alpha Loss_{avis}  + \beta  Loss_{alan}
\end{equation}
%是不是有小标题更好些
%第二种方法的流程，公式以及优缺点 （motivation？？？）
\subsubsection{Language Reconstruction Loss}
The second reconstruction loss is to directly reconstruct the input query based on the attentive proposal features, as Fig. \ref{langloss} shows. First, the concatenation of the original proposal features  $\widetilde { r } _ { s } ^ { i }$, $\widetilde { r } _ { l } ^ { i }$ and $\widetilde { r } _ { c } ^ { i }$ are fed into a one-layer perceptron.
%需要补充一个相同设置下有relu层的实验
\begin{equation}
r_{vis}^i=\phi _ {\rm R e L U } \left(W_v( \left[ \widetilde { r } _ { s } ^ { i }, \widetilde { r } _ { l } ^ { i }, \widetilde { r } _ { c } ^ { i } \right])+b_v  \right)
\end{equation}
Then we calculate the weighted sum of the proposal features according to the total score.
\begin{equation}
{ f } _ { vis } = \sum _ { i = 1 } ^ { N } S _ { t } ^ { i }  { r } _ { vis } ^ { i }
\end{equation}
Based on the fused proposal features, query are generated through LSTM.
\begin{equation}
P ( q | f_{vis} ) = f _ {\rm L S T M } \left(f_{vis} \right)
\end{equation}
We use the same language reconstruction loss as Eq. \ref{lanloss}.
\begin{equation}
Loss _ { lan } = - \frac { 1 } { B } \sum _ { b = 1 } ^ { B } \log ( P ( \hat { q } | f_{vis} ) )
\end{equation}
%直接重建回原来的phrase，不会丢失信息，并且不用考虑视觉重建和语言重建之间的占比。不用考虑视觉重建和语言重建之间的比例并且可以保持网络不会自动学习成0。
Compared to the adaptive reconstruction, the language reconstruction reconstructs the input query directly, so it will not lose any useful language information during training.
%Besides, the second method need not considerate the ration between language and visual reconstruction loss, which is very convenient.
\subsubsection{Attribute Classification Loss}
As mentioned in previous methods \cite{DBLP:conf/iccv/YaoPLQM17,DBLP:journals/pami/WuSWDH18, DBLP:conf/cvpr/Yu0SYLBB18}, attribute information is important on distinguish object of the same category.  Here, we add an attribute classification branch in our model. The attribute label is extracted through an external language parser \cite{DBLP:conf/emnlp/KazemzadehOMB14} according to \cite{DBLP:conf/cvpr/Yu0SYLBB18}.
Subject feature $\widetilde { r } _ { s } ^ { i }$ of proposal is used for attribute classification.
As each query has multiple attribute labels, we use the binary cross-entropy loss for the multi-label classification.

\begin{equation}
Loss _ { a t t  } = f_{\rm BCE} \left( y _ { i j }, p_{ij} \right)
\end{equation}
We use the reciprocal of the frequency that attribute labels appears as weights in this loss to ease unbalanced data.
%如果加入location classification loss在这样子写
%\subsection{Attribute and Location Classification Loss}
%\subsubsection{Attribure Classification}
%\subsubsection{Location Classification}

\subsection{Training and Inference}
The referring expression feature extraction network, the grounding network and the reconstruction network are trained with end-to-end strategy. During training, only query with attribute words goes through attribute classification branch. At inference, the reconstruction module is not needed anymore. We feed the image and query into the network, and get the most related proposal whose final score is the maximal in the grounding module.
\begin{equation}
j = \arg \max _ { i } f  \left( p , r _ { i } \right)
\end{equation}

The final collaborative reconstruction loss is:
\begin{equation}
Loss = Loss_{adp} + \gamma Loss_{lan} +\lambda Loss_{att}
\end{equation}

\section{Experiments}

\subsection{Datasets}
We evaluate our method on four popular benchmarks of referring expression grounding.\\

% 每张图包含同类物体的个数。

\noindent \textbf{RefCOCO \cite{DBLP:conf/eccv/YuPYBB16}.} 
%It is also called UNC RefExp.
The dataset contains 142,209 queries for 50,000 objects in 19,994 images from MSCOCO \cite{DBLP:conf/eccv/LinMBHPRDZ14}.
% It is collected through ReferitGame \cite{DBLP:conf/emnlp/KazemzadehOMB14}.
The dataset is split into train, validation, Test A, and Test B, which has 16,994, 1,500, 750 and 750 images, respectively. Test A contains multiple people while Test B contains multiple objects. Each image contains at least 2 objects of the same object category. 
%The average length of the queries in this dataset is 3.61.\\

\noindent \textbf{RefCOCO+ \cite{DBLP:conf/eccv/YuPYBB16}.} 
It has 141,564 queries for 49,856 referents in 19,992 images from MSCOCO \cite{DBLP:conf/eccv/LinMBHPRDZ14}. 
%It is also collected through ReferitGame \cite{DBLP:conf/emnlp/KazemzadehOMB14}. 
Different from RefCOCO, the queries in this dataset are disallowed to use locations to describe the referents. % Thus, this dataset focus on the appearance of the referents.
The split is 16,992, 1,500, 750 and 750 images for train, validation, Test A, and Test B respectively. Each image contains 2 or more objects of the same object category in this dataset. %The average length of the queries in this dataset is 3.53.\\

\noindent \textbf{RefCOCOg \cite{DBLP:conf/cvpr/MaoHTCY016}.} %It is also called Google Refexp. 
It has 95,010 queries for 49,822 objects in 25,799 images from MSCOCO \cite{DBLP:conf/eccv/LinMBHPRDZ14}. %This dataset is collected in a non-interactive setting on Amazon Mechanical Turk. Thus 
It has longer queries containing appearance and location to describe the referents. 
The split is 21,149 and 4,650 images for training and validation. There is no open test split for RefCOCOg. Images were selected to contain between 2 and 4 objects of the same category. % The average length of the queries is 8.43.\\

\noindent \textbf{RefCLEF \cite{DBLP:conf/emnlp/KazemzadehOMB14}.} %It is also called ReferIt. 
It contains 20,000 annotated images from IAPR TC-12 dataset \cite{article} and SAIAPR-12 dataset \cite{DBLP:journals/cviu/EscalanteHGLMMSPG10}. The dataset includes some ambiguous queries, such as anywhere. It also has some mistakenly annotated image regions. The dataset is split into 9,000, 1,000 and 10,000 images for training, validation and test for fair comparison with \cite{DBLP:conf/eccv/RohrbachRHDS16}. 100 bounding box proposals \cite{DBLP:conf/cvpr/HuXRFSD16} are provided for each image using Edge Boxes \cite{DBLP:conf/eccv/ZitnickD14}. Images contain between 2 and 4 objects of the same object category. The maximum length of all the queries is 19 words.

% 实验部分还需要有， 参数设置， 评价标准， 实验结果， ablation实验
%

\subsection{Experimental Setup}
\subsubsection{Implementation details}
%介绍实验设置的细节，一些超参数的设置，ablation study部分的参数可以不在此介绍
%介绍每个数据集proposals的来源以及个数，在随机的情况下可以达到的results？？？
The proposed ARN is trained through Adam \cite{DBLP:journals/corr/KingmaB14} algorithm with an initial learning rate 4e-4, which is dropped by 10 after every 8,000 iterations. The training iterations are up to 30,000 with a batch size of a single image. Each image has an indefinite number of annotated queries. 
%The rectified linear unit (ReLU) \cite{DBLP:conf/nips/MontufarPCB14} is used as the non-linear activation function. Batch normalization operations are not used in our framework. 
ResNet is our main feature extractor for RoI visual features. 
%\textbf{However, for fair comparison with previous works, the features extracted through VGG network \cite{DBLP:journals/corr/SimonyanZ14a} are also used in our work.}
We adopt EdgeBoxe \cite{DBLP:conf/eccv/ZitnickD14} to generate 100 region proposals for RefCLEF dataset for fair comparison with \cite{DBLP:conf/eccv/RohrbachRHDS16, DBLP:conf/cvpr/ChenGN18}. 
Besides, we also show the performance based on detected objects from Faster R-CNN. 
It is worth noting that we do not extract the context features for RefCLEF dataset. As there are 100 region proposals in each image of the dataset, it is not reasonable to choose one from 5 surrounding proposals as context of the candidate proposal.

% Same as [34], we select 100 proposals produced by proposal generation systems (N = 100).   
% Therefore, we also evaluated with the SSD-detected bounding boxes [20] on the four datasets provided by [46]. 

\begin{table*}[]
	\centering
	\caption{Accuracy (IoU $>$ 0.5) on RefCOCO dataset. \textbf{Bond}: best result. \redfont{Red}: second best result. \bluefont{Blue}: best result of VC. }
	\begin{tabular}{c|c|ccc|ccc|c}
		\hline
		\multirow{2}{*}{Methods}&\multirow{2}{*}{Settings}&\multicolumn{3}{c|}{RefCOCO}&\multicolumn{3}{c|}{RefCOCO+}& RefCOCOg\\ 
		\cline{3-9}&&val&testA&testB&val&testA&testB& val\\ \hline 
		VC  & w/o reg& - &13.59& 21.65& - & 18.79 & 24.14 & 25.14 \\ 
		VC & -& - & 17.34 & 20.98 & - & 23.24 & 24.91 & \bluefont{33.79} \\ 
		VC  & w/o $\alpha$ & - & \bluefont{33.29} & \bluefont{30.13} & - & \bluefont{34.60} & \bluefont{31.58} & 30.26 \\ \hline
		VC  (det) & w/o reg & - & 17.14 & 22.30 & - & 19.74 & 24.05 & 28.14 \\ 
		VC (det) &- & - & 20.91 & 21.77 & - & 25.79 & 25.54 & 33.66 \\ \
		VC (det) & w/o $\alpha$  & - & 32.68 & 27.22 & - & 34.68 & 28.10 & 29.65 \\ \hline \hline
		
		ARN & $L_{adp}+L_{att}$ & 33.07 & \textbf{36.43} & 29.09 & 33.53 & \textbf{36.40} & 29.23 & 33.19 \\
		
		ARN &  $L_{lan}+L_{att}$& \textbf{38.05} & 35.27& \textbf{36.47 }& \redfont{34.51} & 34.40 & \textbf{36.12} & \textbf{39.62}\\
		
		ARN & $L_{lan}+L_{adp}$ & 33.60 & 35.65& 31.48 & 34.40 & 35.54 & 32.60 & 34.50\\
		
		ARN (det)& $L_{lan}+L_{adp}$ & 31.58 & 35.50& 28.32 & 31.73 & 34.23 & 29.35 & 32.60\\
		
		ARN & $L_{lan}+L_{adp}+L_{att}$  &\redfont{34.26}&\redfont{36.01}&\redfont{33.07}&\textbf{34.53}&\redfont{36.01}&\redfont{33.75}&\redfont{34.66}\\ 
		ARN (det) & $L_{lan}+L_{adp}+L_{att}$  &32.17&35.35&30.28&32.78&34.35&32.13&33.09\\\hline
		
	\end{tabular}
	\label{refcoco}
\end{table*}
\begin{table*}[]
	\centering
	\caption{Albation study on RefCOCO dataset.}
	\scalebox{1.1}{
		\begin{tabular}{c|cccc|ccc|ccc|c}
			\hline
			\multirow{2}{*}{}&\multicolumn{4}{c|}{Settings}&\multicolumn{3}{c|}{RefCOCO}&\multicolumn{3}{c|}{RefCOCO+}& RefCOCOg\\ 
			\cline{2-12}&$\alpha$&$\beta$&$\gamma$&$\lambda$&val&testA&testB&val&testA&testB&val\\ \hline 
			
			case 1  &  1&1  &1  &0   &32.92  & 36.40 &29.26&33.06&36.34&29.60&33.08 \\
			
			case 2 &   0.01&1  &1  &1   &34.32 &36.24&33.05&35.60&36.92&33.09&34.44 \\
			
			case 3 & 0.01&1&5&1&34.26&36.01&33.07&34.53&36.01&33.75&34.66\\
			case 4& 0.01&1&10&1&34.18&35.83&32.29&32.39&33.39&32.89&34.24\\
			case 5& 0.01&1&15&1&29.09&27.13&33.09&29.97&27.98&33.99&34.94\\
			case 6&0.01&1&20&1&29.87&27.86&33.05&29.20&25.57&35.28&34.60\\
			%exp7 (34)& 0.01&1&1&1&0&&&&&&&\\

			\hline
			
	\end{tabular}}
	
	\label{refcoco_ablation}
\end{table*}
\subsubsection{Metrics}
%介绍评价标准IoU
The Intersection over Union (IoU) between the selected region and the ground-truth are calculated to evaluate the network performance. If the IoU score is greater than 0.5, the predicted region is considered as the right grounding.

\subsection{Results}

\begin{figure*}[tp]
	\centering
	·	\includegraphics[width=0.85\textwidth]{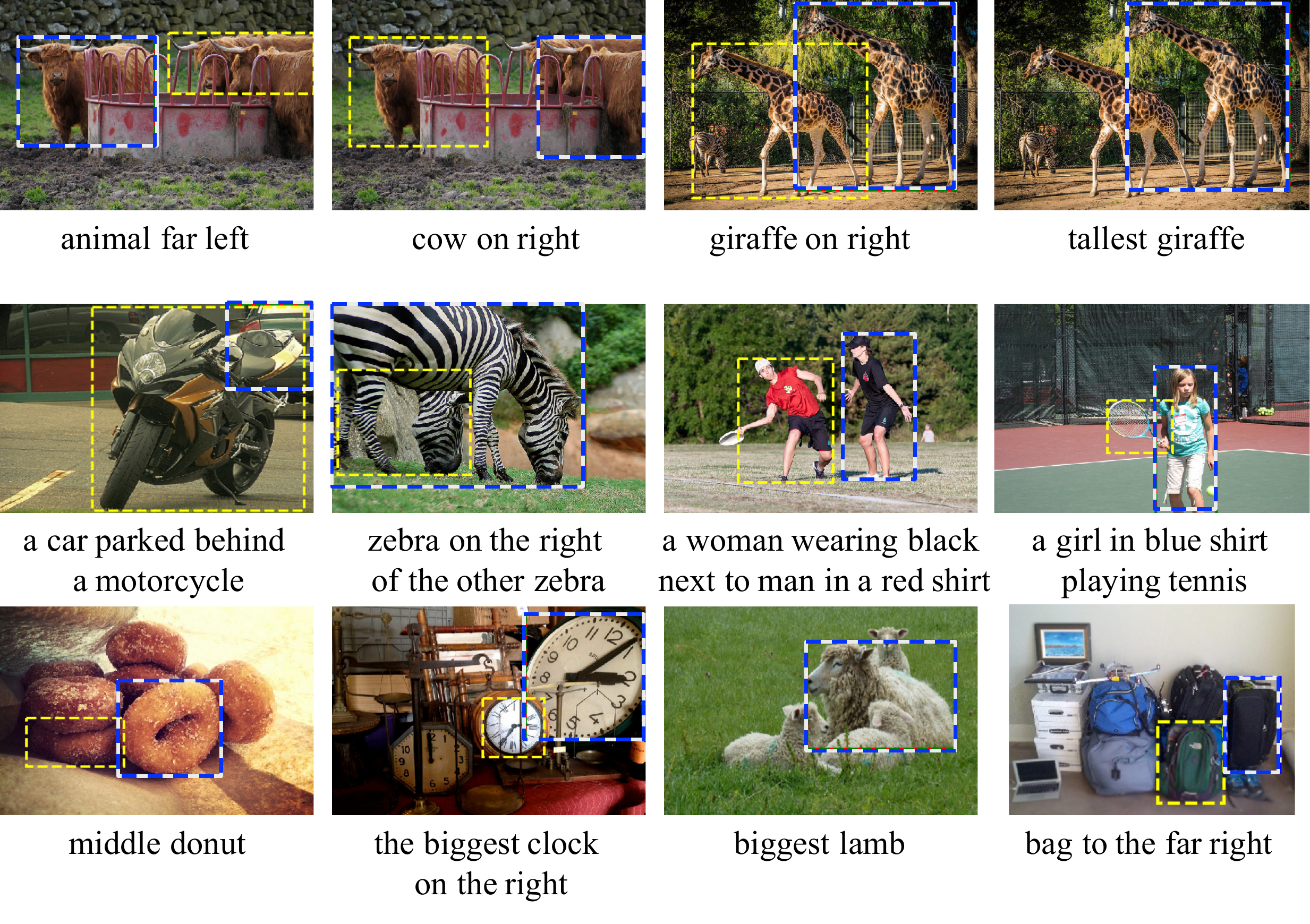}
	\caption{Qualitative results on MSCOCO datasets. The denotations of the bounding box colors are as follows. Solid white: ground truth; dashed blue: predicted proposal; dashed yellow: context ground. }
	\label{visualization}
	
\end{figure*}

\subsubsection{RefCOCO Datasets}
\textbf{Performance Analysis:}
Table \ref{refcoco} reports the results on RefCOCO, RefCOCO+ and RefCOCOg datasets. We compared the proposed ARN with the only published unsupervised results on these datasets \cite{DBLP:conf/cvpr/ZhangNC18}. 
We can have the following findings. First, adaptive reconstruction performs better on testA, which contains multiple people. Language reconstruction achieves better performance on testB, which contains multiple other objects. Second, the collaborative loss can get second best on all the test, indicating it can better handle different kinds of datasets. %模型可扩展的能力   
We also show the results using detected object proposals from Faster R-CNN. The performance drops due to detection error.\\

\noindent \textbf{Ablastion Study:}
Table \ref{refcoco_ablation} reports the results on RefCOCO datasets with different settings. $\alpha$, $\beta$, $\gamma$, $\lambda$ denoted the weights on $Loss_{avis}$, $Loss_{alan}$, $Loss_{lan}$, $Loss_{att}$, respectively.  %$\omega_{adp}$ denotes whether to use the adaptive grounding. If $\omega_{adp}$ is 1, it means that ARN assign different weights on subject, location and context scores according to different expressions. If  $\omega_{adp}$ is 0, ARN assign the same weightiness on subject, location and context.
The proportion is based on the order of magnitude of different losses. 
We find that when $Loss_{lan}$ accounts for a more significant part in the collaborative loss, the performance on testA will drop greatly. While when the proportion of $Loss_{adp}$ is bigger, the results in testB will be a disaster. After the parameter search, we find that the settings in case5 get good result on all the datasets.

%首先是最基本的设置比较，在VGG特征下是多少，ResNet101特征下是多少。分成两个表格：RefCOCO系列是一个表，仅有一个对比方法， RefClef是一个表，有多个对比方法
% 实验部分可以说明一下external language parser并不能对效果有显著提升， However, conventional parsers (\eg, Standford Dependency) are observed to be suboptimal to the visual grounding task \cite{DBLP:conf/cvpr/HuRADS17}.We can see that there is no significant improvement of VC w/ parser over VC w/o α.\cite{DBLP:conf/cvpr/ZhangNC18} the formal representations of language produced by syntactic parsers do not always correspond to intuitive visual representations. \cite{DBLP:conf/acl/ZhuZCZZ13}
\begin{table}[]
	\centering
	\caption{Accuracy (IoU $>$ 0.5) on RefCLEF dataset.}
	\scalebox{0.9}{	\begin{tabular}{l|l}
			\hline
			Method  & IoU  \\ \hline
			LRCN \cite{DBLP:conf/cvpr/DonahueHGRVDS15}   & 8.59     \\ 
			Caffe-7K \cite{DBLP:conf/rss/GuadarramaRSZFD14} & 10.38    \\ 
			GroundeR \cite{DBLP:conf/eccv/RohrbachRHDS16}& 10.70    \\ 
			MATN  \cite{DBLP:conf/cvpr/0006LZF18} & 13.61    \\ 
			VC \cite{DBLP:conf/cvpr/ZhangNC18} & 14.11\\
			VC w/o $\alpha$ \cite{DBLP:conf/cvpr/ZhangNC18} & 14.50\\
			KAC Net \cite{DBLP:conf/cvpr/ChenGN18} & 15.83    \\ \hline \hline
			ARN ($loss_{lan}$) & 21.86 \\
			ARN ($loss_{lan}+loss_{adp}$)& 25.35 \\
			ARN ($loss_{lan}+loss_{adp}+loss_{att}$)& \textbf{26.19} \\ \hline
	\end{tabular}}
	\label{refclef}
\end{table}

\subsubsection{RefCLEF Dataset}
\textbf{Performance Analysis:}
We compare our adaptive reconstruction network (ARN) with state-of-the-art supervised referring expression grounding methods. 
%The LRCN use the image captioning to score how likely the query phrase is to be generated for the proposal box. CAFFE-7K predicts a class for each proposal box and then compared to the query phrase after both are projected to a joint vector space. 
%GroundeR learns to ground by reconstructing a given phrase using an attention mechanism. KAC Net leverage complementary external knowledge to ground the referent.
Table \ref{refclef} reports the results on RefCLEF dataset. We can see that ARN outperforms state-of-the-art result by 10.36\%. We can have the following observations. 
First, with only language reconstruction loss, our method outperforms state-of-the-art result by 6.03\%, which indicates our proposed adaptive grounding module taking effect. Second, adding our proposed adaptive reconstruction module, the performance achieves another 3.49\% increase compared to with language reconstruction loss only. Third, the attribute classification loss also helps localization, the performance increase by 0.84\% compared to previous result. \\

\noindent \textbf{Ablation Study:}
We study the benefits of each loss module by running ablation experiments. Table \ref{refclef_ablation} reports the results on RefCLEF dataset with different loss proportion. $\alpha$, $\beta$, $\gamma$, $lambda$ denoted the weights on $Loss_{avis}$, $Loss_{alan}$, $Loss_{lan}$, $Loss_{att}$, respectively. The adaptive visual reconstruction loss is first set as 0.001 based on the order of magnitude. We can have the follow ablation experiment. We change the proportion of $Loss_{avis}$ and $Loss_{alan}$ in case 2 and case 3 compared to case 1, respectively. We find the result is better when $\alpha$ is 0.001, due to the order of magnitude in $Loss_{avis}$. The comparison of case 1 and case 6, case 4 and case 5 show that attribute classification loss can improve the grounding results. case 6, case 7 ,case 8 and case 9 show that when the proportion of $Loss_{lan}$ is 30, the performance of the network will be better. However, when we only use the $Loss_{lan}$ in case 10, the results are not as good as the combination of $Loss_{adp}$ and $Loss_{lan}$.

\begin{table}[]
	\centering
	\caption{Ablation study on RefCLEF dataset.}
	\scalebox{1}{\begin{tabular}{c|cccc|c}
			\hline
			  & $\alpha$&$\beta$&$\gamma$&$\lambda$&val \\ \hline
			%exp1 & 1 &1&1&0&23.03\\
			case 1 & 0.001&1&10&0&24.14\\
			case 2 & 0.01 &1 &10&0&21.83\\ 
			case 3 & 0.001 &10 &10&0&22.55\\
			\hline
			case 4& 0.001&1&1&0&22.34\\ 
			case 5& 0.001&1&1&1&25.35\\ \hline
			
			case 6 & 0.001&1&10&1&24.34\\ 
			case 7& 0.001&1&20&1&24.76\\ 
			case 8& 0.001&1&30&1&26.19\\
			case 9&0.001&1&40&1&25.53\\
			\hline
			case 10&0&0&1&0&21.86\\

			\hline
	\end{tabular}}
	
	\label{refclef_ablation}
\end{table}

\subsubsection{Qualitative Results}

Fig. \ref{visualization} shows qualitative example predictions on RefCOCO, RefCOCO+ and RefCOCOg datasets. The query is shown below corresponding images. The ground truth, grounding proposal and context proposal are denoted as solid white, dashed blue and dashed yellow, respectively. The first row shows the result based on different query in the same image. The proposed ARN can handle the location information correctly. The second row shows some examples with context information. ARN correctly grounds both the referential object and context object. The third row shows some difficult examples where multiple objects of the same category exist. It shows that the proposed ARN can help to ground in the hard cases which contain multiple objects of the same category.

\section{Conclusion}
To address the weakly supervised referring expression grounding problem, we propose a novel end-to-end adaptive reconstruction network. The ARN models the mapping between image proposal and query upon the subject, location and context information through adaptive grounding and collaborative reconstruction. 
Specially, a hierarchical attention model is designed to adaptively ground the query on the proposal with proposal attention and language attention.
This model is trained by minimizing a collaborative reconstruction loss, which consists of language reconstruction loss, adaptive reconstruction loss and attribute classification loss. Experiments demonstrate that the proposed method provides a significant improvement in performance on RefCLEF, RefCOCO, RefCOCO+ and RefCOCOg datasets. 
%We find that it is hard to find the correct context region in more complex scenario as there is neither ground truth for the referential object nor context object. We will make more effort on this in our future work.

\noindent\textbf{Acknowledgements.} This work was supported in part by National Natural Science Foundation of China: 61771457, 61732007, 61772494, 61672497, 61622211, 61836002, 61472389, 61620106009 and U1636214, in part by Key Research Program of Frontier Sciences, CAS: QYZDJ-SSW-SYS013, and Fundamental Research Funds for the Central Universities under Grant WK2100100030.

{\small
\bibliographystyle{ieee_fullname}
\bibliography{egbib}
}

\end{document}